\documentclass[journal]{IEEEtran}
\usepackage{multirow} 
\usepackage{amsmath,amsfonts}
\usepackage{algorithmic}
\usepackage{algorithm}
\usepackage{array}
\usepackage{booktabs}
\usepackage[caption=false,font=normalsize,labelfont=sf,textfont=sf]{subfig}
\usepackage{textcomp}
\usepackage{stfloats}
\usepackage{url}
\usepackage{verbatim}
\usepackage{graphicx}
\usepackage{cite}
\usepackage{amsmath}
\usepackage[table,xcdraw]{xcolor}
\usepackage{amssymb}
\usepackage{xcolor}
\usepackage{amsfonts}
\definecolor{rblue}{rgb}{0,0.5,1}
\definecolor{hollywoodcerise}{rgb}{0.96, 0.0, 0.63}
\definecolor{lasallegreen}{rgb}{0.03, 0.47, 0.19}
\definecolor{hanpurple}{rgb}{0.32, 0.09, 0.98}
\definecolor{green(pigment)}{rgb}{0.0, 0.65, 0.31}
\usepackage[pagebackref=false,breaklinks=true,colorlinks,bookmarks=false]{hyperref}
\hypersetup{colorlinks,linkcolor={red},citecolor={hanpurple},urlcolor={magenta}}

\DeclareUnicodeCharacter{2061}{}
\begin{document}

\title{GenMapping:\\Unleashing the Potential of Inverse Perspective Mapping for Robust Online HD Map Construction}

\author{Siyu Li$^{1}$, Kailun Yang$^{1}$, Hao Shi$^{2,4}$, Song Wang$^{3}$, You Yao$^{5}$, and Zhiyong Li$^{1}$
\thanks{This work was supported in part by the National Natural Science Foundation of China (No. U21A20518, No. U23A20341, and No. 62473139) and in part by Hangzhou SurImage Technology Company Ltd. \textit{(Corresponding authors: Kailun Yang and Zhiyong Li.)}}
\thanks{$^{1}$S. Li, K. Yang, and Z. Li are with the School of Robotics and the National Engineering Research Center of Robot Visual Perception and Control Technology, Hunan University, Changsha 410082, China (email: kailun.yang@hnu.edu.cn; zhiyong.li@hnu.edu.cn).}%
\thanks{$^{2}$H. Shi is with the State Key Laboratory of Extreme Photonics and Instrumentation, Zhejiang University, Hangzhou 310027, China.}%
\thanks{$^{3}$S. Wang is with the College of Computer Science, Zhejiang University, Hangzhou 310027, China.}%
\thanks{$^{4}$H. Shi is also with Shanghai Supremind Technology Company Ltd, Shanghai 201210, China.}
\thanks{$^{5}$Y. Yao is with the USC Viterbi School of Engineering, the University of Southern California, Los Angeles 90089, California, United States.}
}%

\maketitle

\begin{abstract}
Online High-Definition (HD) maps have emerged as the preferred option for autonomous driving, overshadowing the counterpart offline HD maps due to flexible update capability and lower maintenance costs. However, contemporary online HD map models embed parameters of visual sensors into training, resulting in a significant decrease in generalization performance when applied to visual sensors with different parameters. Inspired by the inherent potential of Inverse Perspective Mapping (IPM), where camera parameters are decoupled from the training process, we have designed a universal map generation framework, GenMapping. The framework is established with a triadic synergy architecture, including principal and dual auxiliary branches. When faced with a coarse road image with local distortion translated via IPM, the principal branch learns robust global features under the state space models. The two auxiliary branches are a dense perspective branch and a sparse prior branch. The former exploits the correlation information between static and moving objects, whereas the latter introduces the prior knowledge of OpenStreetMap (OSM). The triple-enhanced merging module is crafted to synergistically integrate the unique spatial features from all three branches. To further improve generalization capabilities, a Cross-View Map Learning (CVML) scheme is leveraged to realize joint learning within the common space. Additionally, a Bidirectional Data Augmentation (BiDA) module is introduced to mitigate reliance on datasets concurrently. A thorough array of experimental results shows that the proposed model surpasses current state-of-the-art methods in both semantic mapping and vectorized mapping, while also maintaining a rapid inference speed. Moreover, in cross-dataset experiments, the generalization of semantic mapping is improved by $17.3\%$ in mIoU, while vectorized mapping is improved by $12.1\%$ in mAP. The source code will be publicly available at \url{https://github.com/lynn-yu/GenMapping}.

\end{abstract}


\section{Introduction}
\IEEEPARstart{O}{nline} High-Definition (HD) map models, benefiting from flexible mapping and lower costs, have recently achieved significant breakthroughs~\cite{bevsurvey,mapsurvey}. 
Currently, HD maps are categorized into two types: semantic mapping and vectorized mapping. Semantic mapping, describing road areas in a grid format, is extensively used in end-to-end autonomous driving models~\cite{uniad,vad,genad}.
Vectorized mapping represents road instances with points and lines, which is lightweight and better suited for path planning and prediction tasks~\cite{densetnt,hivt}.

\begin{figure}[t!]
      \centering
      \includegraphics[scale=0.30]{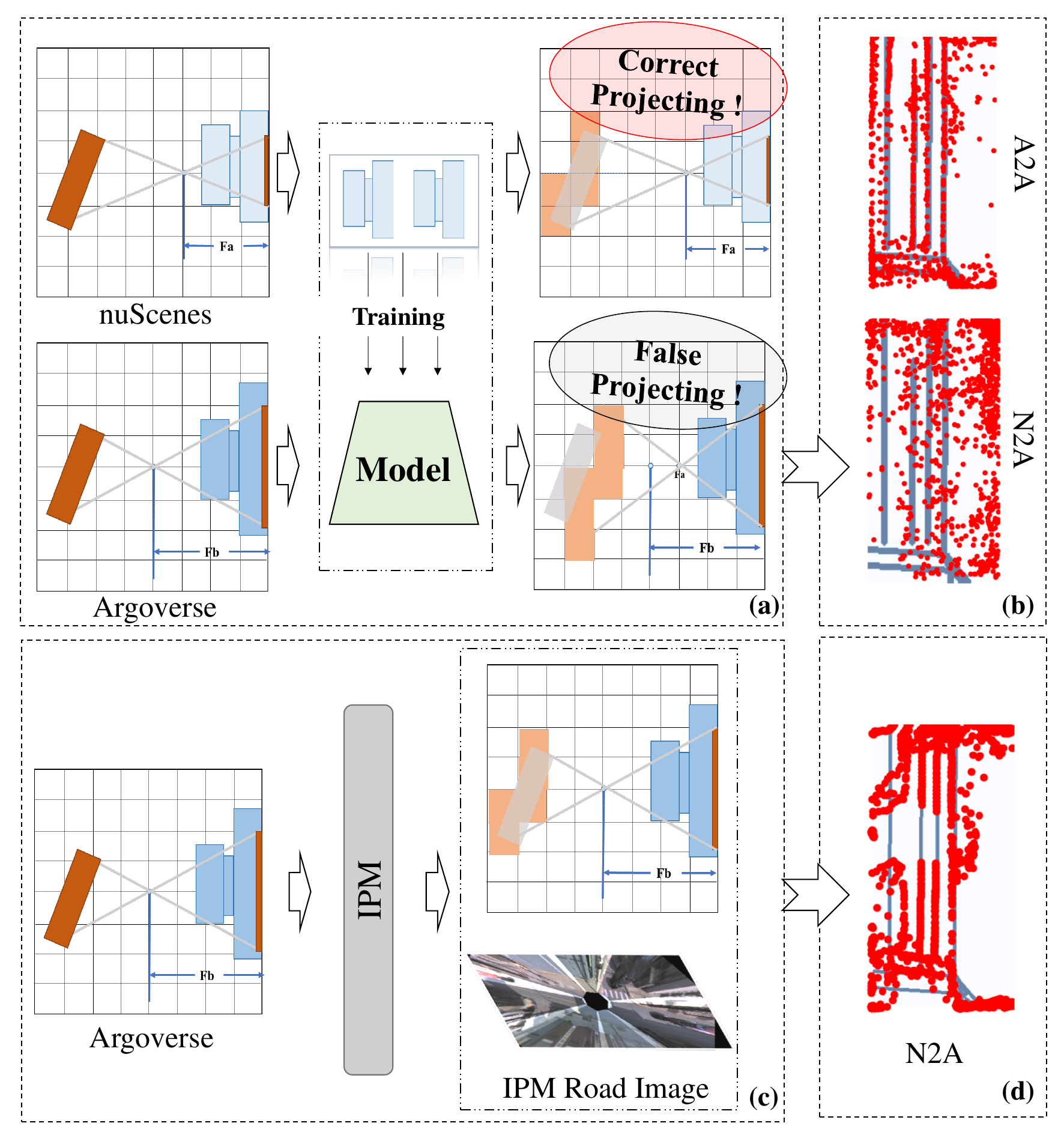}
      \caption{Generalization analysis of HD mapping models facing cross-dataset shift. `N2A' denotes the validation result of the model trained on the nuScenes dataset~\cite{nus} evaluated on Argoverse~\cite{argoverse}. `A2A' follows the same definition. (a) and (b) represent the cross-dataset performance of a state-of-the-art mapping method~\cite{maptrv2}. Inconsistent sensor parameters between training and validation lead to projection errors, causing inaccurate detection of the positions of map instances. 
      (c) and (d) illustrate the cross-dataset results based on the proposed method, leveraging the advantage of decoupling the sensor parameters.}
      \label{fig.intro_1}
      \vspace{-1.5em}
\end{figure}

\begin{figure*}[t!]
      \centering
      \includegraphics[scale=0.40]{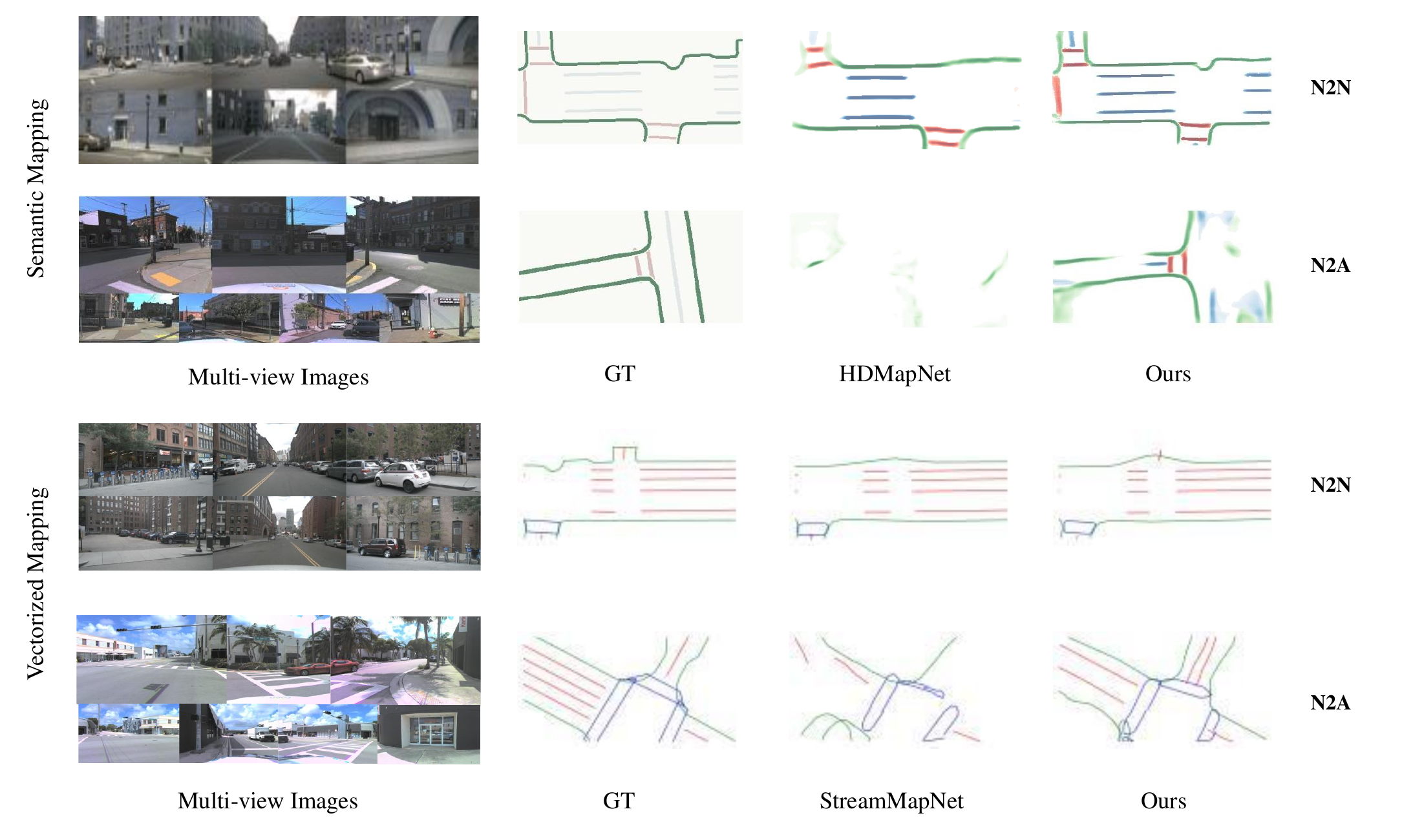}
      \vspace{-0.55cm}
      \caption{Mapping accuracy and generalization performance of HD map models on public datasets. The first two rows depict semantic mapping results and the last two rows depict vectorized mapping results. The figure shows the visualization results of the model trained on nuScenes~\cite{nus} on the validation sets of nuScenes (N2N) and Argoverse~\cite{argoverse} (N2A), respectively. The proposed method adopts a triadic synergy framework established with the concept of parameter decoupling, leading to stronger generalization performance.
      }
      \label{fig.intro_2}
      \vspace{-1.5em}
\end{figure*}

HD maps are constructed in the Bird's Eye View (BEV) where the coordinate system is perpendicular to the perspective view. 
If the vision sensor parameters and depth values are available, converting the perspective features to the BEV space becomes straightforward. 
The challenge appears when accurate depth values are not available, which are often difficult to measure in real-world driving scenes.
Therefore, view transformation methods are focused on studying visual HD maps.
The view transformation of HDMapNet~\cite{hdmapnet} implicitly learned intrinsic parameters and depth through Multi-Layer Perceptron (MLP)~\cite{mlp} layers. MapTRv2~\cite{maptrv2} designed a depth estimation network embedded with the intrinsic and extrinsic parameters referenced from the dataset. These methods project perspective features to BEV space based on depth values and camera parameters, referred to as 2D-to-3D. In contrast, StreamMapNet~\cite{streammapnet} adopted the 3D-to-2D transformation, where 3D point features obtained by projection relations with the visual features were compressed to BEV features from a height space.
Although these ingenious designs exhibit remarkable performance on a single dataset, they are prone to overfitting and failing to operate effectively in environments with different sensor configurations, as these models incorporate visual sensor parameters into the model training. 

As illustrated in Fig.~\ref{fig.intro_1}(a), a set of generalization analyses of cross-dataset performance for a depth-based method~\cite{maptrv2} evidences the severe performance degradation issue. 
The absolute depth estimation of visual images is closely related to camera parameters. 
When a map model trained on camera A (\textit{e.g.}, on nuScenes) is applied to camera B (\textit{e.g.}, on Argoverse),  the network typically uses the camera parameters from camera A to estimate depth. 
Even with the camera parameter integrated into the model training, the generalization performance remains unsatisfactory, struggling to learn the correct map structure, as shown in Fig.~\ref{fig.intro_1}(b).
Thus, we ask whether, decoupling the visual sensor parameters from the training process, could benefit the generalization. 
Inverse Perspective Mapping (IPM) technology with powerful prior knowledge for road structures comes to our attention~\cite{ipm-1,gafb}. 
IPM, a special case of the 3D-to-2D mode, sets 3D points at a fixed height to obtain BEV road images that are the learned objects of a map model. 
Naturally, visual sensor parameters are decoupled from model learning, which is advantageous for the deployment across data domains. 
Nevertheless, as shown in Fig.~\ref{fig.intro_1}(c), IPM images suffer from data distortion and lack context interaction above the road plane which is important in BEV understanding~\cite{ci3d}.

To unleash the powerful generalization capabilities of IPM and address the above challenges, we propose a universal online HD map construction model, GenMapping.
The framework is established with a triadic synergy architecture, including principal and dual auxiliary branches.
Due to the local geometric distortions presented in IPM images, the principal branch introduces a module based on the State Space Model (SSM)~\cite{ssm} to mitigate these local distortion problems. 
The dense perspective auxiliary branch learns dense associations between dynamic and static objects within the perspective coordinate system. 
The sparse prior auxiliary branch encodes drivable areas implicitly based on OSM~\cite{openstreetmap} describing the road centerline with vectorized lines. 
In addition, a triple-enhanced merging module is designed and embedded into the principal branch, integrating auxiliary features through progressively layered fusion.
At the same time, joint learning and data augmentation methods are presented to improve generalization ability. 
On the one hand, a Cross-View Map Learning (CVML) module is proposed creatively under a mutual constraint space between the perspective view and BEV. 
On the other hand, facing aligned features in different spaces, Bidirectional Data Augmentation (BiDA) is designed to reduce the dependence on the training dataset. 
As verified in Fig.~\ref{fig.intro_2}, GenMapping achieves outstanding performance on the public nuScenes dataset~\cite{nus}. 
Furthermore, experiments facing cross-dataset transfer, \textit{i.e.}, shifting from nuScenes (N) to Argoverse~\cite{argoverse} (A), demonstrate the superiority of the proposed method in robust online HD map construction against other state-of-the-art approaches.
The main contributions of this work are summarized as follows:
\begin{itemize}
    \item We introduce an accurate and robust HD map model, GenMapping. It is a triadic framework centered around IPM. Mitigating local distortion issues through a sequence learning mechanism, while employing triple-enhanced merging to address the sparsity of IPM images.
    \item We propose the Cross-View Map Learning (CVML) module for the mutual constraints between perspective and BEV space to strengthen the robustness of the model from the joint learning level. 
    \item We design the Bidirectional Bata Augmentation (BiDA) component to enhance model generalization. It is a plug-and-play module that can be seamlessly integrated into other tasks and consistently improve generalizability.
    \item Extensive experiments demonstrate the superiority of the proposed method and the strong generalization across different HD map construction scenarios.
\end{itemize}

\section{Related Work}
In this section, we present related works in three parts: view transformation for BEV understanding, as well as recent advances in HD maps and state space models.

\subsection{View Transformation for BEV Understanding}
Since monocular cameras without depth information are the main focus of current research, the transformation between visual perspective and BEV coordinate systems is challenging.
According to the research of view transformation, the current methods can be roughly divided into two categories: 2D-3D~\cite{LSS,vpn,bevdepth,divide} and 3D-2D transformation methods~\cite{BEVFormer,ipm-1}. 

Depth, an important factor for BEV understanding, is the crux of 2D-3D methods. 
Based on a depth estimation network, LSS~\cite{LSS} combined the intrinsic and extrinsic parameters to project the perspective features to obtain BEV features. 
To enhance the robustness of depth estimation, BEVDepth~\cite{bevdepth} supervised the model through ground truth of depth from LiDAR sensors. 
Different from the aforementioned works, VPN~\cite{vpn} learned depth and camera parameters simultaneously via Multi-Layer Perceptron (MLP) layers. 
Furthermore, PON~\cite{pon} with a similar strategy further studied the relationship between different resolution perspective features and BEV features at different distances.
However, it is crucial to acknowledge that the estimation of absolute depth cannot be separated from the camera parameters, indicating that these methods are excessively reliant on consistent camera parameters within the dataset.

The other type of approach, 3D-2D methods, is to compress 3D features obtained from corresponding 2D perspective features, where the height estimation is of paramount importance. 
In the work of BEVFormer~\cite{BEVFormer, BEVFormerv2}, uniformly distributed 3D points, obtained by equidistant sampling of height values, were projected onto the perspective view to fuse local features based on learnable sampling points. Then, BEV features could be obtained by compressing 3D features of different heights. 
Compared with the former setting which fixes the 3D point coordinates directly, CVT~\cite{cvt} used learnable queries to learn spatial location features implicitly. 
Trans2Map~\cite{trans2map} found the correspondence between the epipolar line of BEV and the columns on the perspective image. 
Thus, a hybrid attention mechanism was chosen to obtain BEV features between the pairs of these lines. 
Typically, given the same camera parameters in a dataset, the height value is relatively easy to learn in a 3D space. 
However, accurately detecting height from images with different parameters is quite challenging. 
Similar to depth estimation, height detection is strongly correlated with camera parameters, illustrating that generalization across different datasets is difficult.
IPM~\cite{ipm-1,ipm-2} is a special case of this category where a fixed height is adopted. Although it is friendly for road plane detection, it has been rarely applied in recent research about HD maps. On the one hand, this is due to the distortion problem caused by uneven road surfaces~\cite{gafb}; on the other hand, it is because of the information sparsity resulting from the lack of interaction with information on the road plane. However, IPM with great abilities for generalization performance deserves to have their potential explored.  

\subsection{Online HD Map Learning}

\subsubsection{Semantic Mapping}
Semantic mapping constructs road maps in a grid format within the BEV space 
HDMapNet~\cite{hdmapnet} is a seminal work that employed an approach through MLP to implicitly learn depth and internal parameters. 
Similarly, BEVSegFormer~\cite{bevsegformer} proposed to neglect the parameters of cameras, where learnable queries were selected to obtain BEV features from the perspective features through the attention mechanism. 
In the work of BEVerse~\cite{beverse}, semantic mapping was an auxiliary task that assisted in constructing a unified framework for perception and prediction.
LSS~\cite{LSS} was chosen as the view transformer module and the temporal fusion module was introduced to improve the accuracy of perception. 
Recently, P-MapNet~\cite{pmapnet} explored the prior performance of OpenStreetMap (OSM) for long-distance semantic mapping. 
Certainly, there are other aspects to explore in improving the quality of mapping, including offline map fusion~\cite{mvmap}, map update~\cite{neruralmap}, and satellite map fusion~\cite{satellite}, which are the driving force for precise structures in an HD map.

\subsubsection{Vectorized Mapping}
Vectorized mapping offers a more lightweight approach, combining points and lines.
Bezier curves are a suitable structure for describing vector objects, which has been applied in early mapping works~\cite{stsu,bezier-1,bezier-2}. 
Nowadays, vectorized mapping tends to be polylines rather than Bezier curves, because of their simpler structure. 
VectorMapNet~\cite{vecmapnet} is the first work to construct vector maps with polylines, where IPM was regarded as the view transformation module. 
Based on the deep perspective features, it adopted IPM to obtain BEV features. 
However, they neglected the importance of learning from raw IPM images, which can potentially learn the original road geometry in the BEV space.
MapTR~\cite{maptr} developed a more streamlined framework for vectorized mapping. GKT~\cite{gkt}, featuring a geometry-guided kernel transformer based on BEVFormer, was chosen as the default view transformer. 
Several decoder layers based on the Deformable DETR~\cite{ddetr} were used in its map decoder. 
In their recent work~\cite{maptrv2}, the impact of a combination of BEVPool ~\cite{bevpoolv2} and depth supervision was explored in vectorized mapping. It is no surprise that this combination produced remarkable results. 
In subsequent research, part of the focus is on the study of map decoders~\cite{mgmap,admap,pivotnet}, as an appropriate map decoder can significantly enhance map accuracy. 
Another aspect of the research~\cite{streammapnet,maptracker} is concentrated on the temporal fusion. StreamMapNet~\cite{streammapnet}, which used the streaming strategy widely applied in object detection~\cite{streampetr,sparse4d}, is a classic solution for large-scale temporal fusion. MapTracker~\cite{maptracker} introduced the concept of target tracking into temporal fusion. 
Since the goal of vectorized mapping is to serve downstream tasks, there is ongoing research work~\cite{accelerating,mapuncertainty} exploring how to seamlessly integrate it into tasks such as path planning. 
Similarly, the ability of a model to be applied flexibly in real-world environments is also crucial, implying that a model with strong generalization and robustness is worth investigating.

However, the generalization of online HD map learning is under-explored, with only the work of SemVecNet~\cite{semvecnet}. 
It projected semantic labels of perspective images to construct a semantic BEV map through the depth value from LiDAR. Then, a semantic BEV map, as an intermediate representation, was translated into a vector map. 
However, this approach requires preprocessing of perspective images and can only be applied in an environment equipped with LiDAR, limiting its applicability in a real environment.
Unlike this work, we focus on establishing a flexible HD map construction framework with camera-only observations that can be applied to various existing architectures of online HD mapping.

\begin{figure*}[tb]
      \centering
      \includegraphics[scale=0.40]{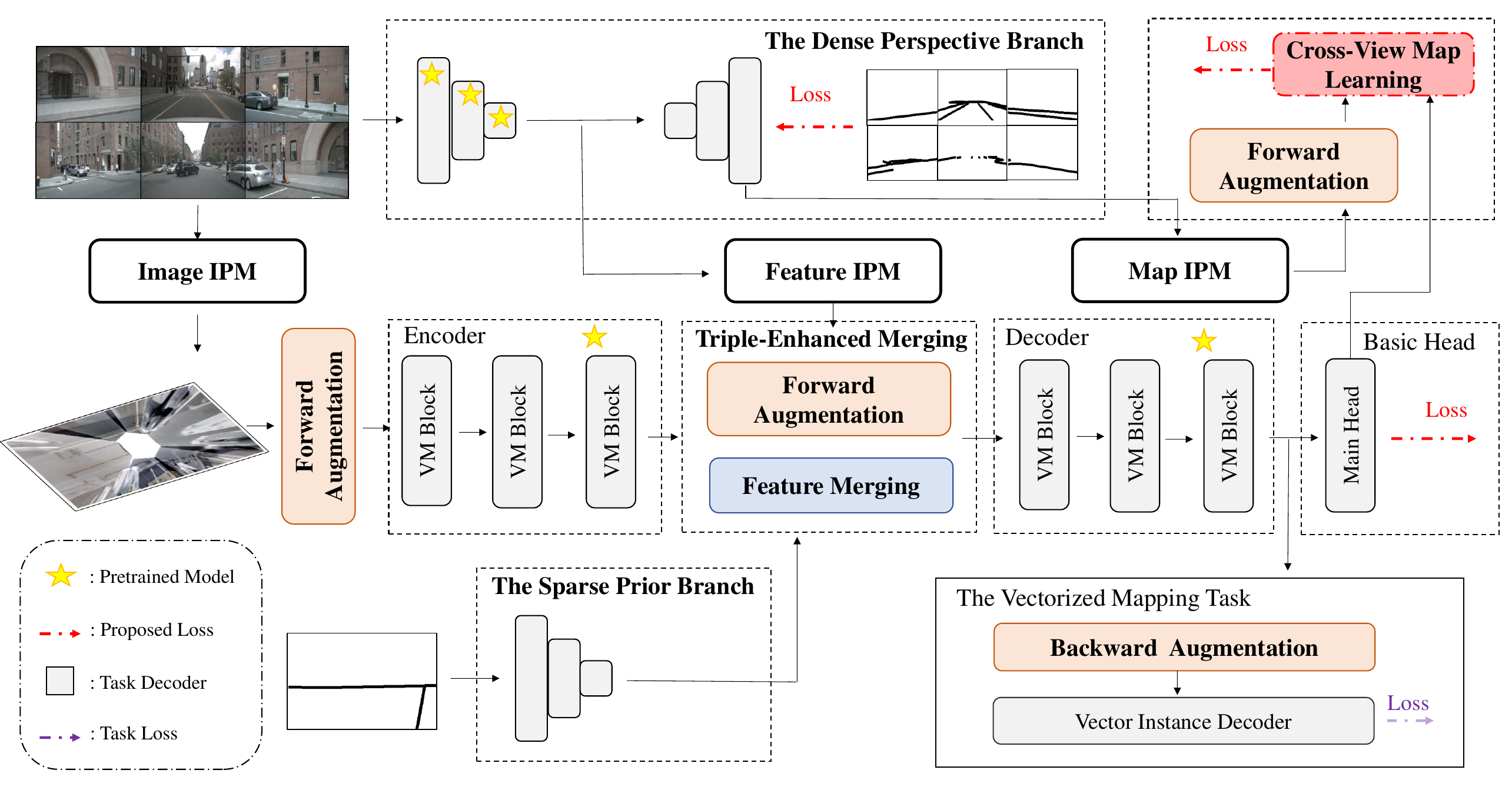}
      \caption{ Overview of the established GenMapping framework for robust online HD map construction.
      The pipeline follows a triadic synergy architecture with principal and dual auxiliary branches. The triple-enhanced merging module synchronously fuses three-way features in BEV space. Bidirectional data augmentation, including forward and backward augmentation, and the cross-view map learning module are designed to enhance mapping robustness.
      }
      \label{fig.frame}
      \vspace{-1.0em}
\end{figure*}
\subsection{State Space Models}
The recently appeared State Space Model (SSM)~\cite{ssm} has been attractive for a wide variety of tasks for establishing long-distance dependencies. 
In particular, Mamba~\cite{mamba} reduced computational complexity, allowing it to shine in research with long sequence data, \textit{i.e.}, language understanding~\cite{ssm-lan}. 
It was also introduced in visual learning, which involved computationally intensive reasoning tasks. 
U-Mamba~\cite{umamba} and SegMamba~\cite{segmamba} mixed Mamba and convolutional neural network structures to achieve semantic scene segmentation. 
VMamba~\cite{vmamba} exhibited linear complexity with the advantages of the global receptive field and dynamic weight. 
It introduced a cross-scan module to merge 1D sequence features, which had a four-way selective scan methodology.
MambaVision~\cite{mambavision} proposed a hybrid framework, which can capture both short and long-range dependencies.
VM-Unet~\cite{vmunet} proposed a segmentation framework for UNet with VMamba as the basic unit. It demonstrated that the SSM-based architecture not only improved the accuracy of semantic segmentation but also reduced the computational complexity, which was an effective SSM-based segmentation baseline.
Inspired by the successful application of these SSM works, we further explore their application to HD maps. In particular, we aim to investigate and materialize the power of SSM to accurately detect road instances in IPM images with geometric distortions. 

\section{Method}

\subsection{Problem Formulation}
\subsubsection{Inverse Perspective Mapping}
The BEV plane is divided into independent small grids, representing $(X_i, Y_i)$ in the ego coordinate system. 
Given multi-view perspective images $I_n$ (or features $F$, maps $M$) with intrinsic $T_{n}^{in}$ and extrinsic $T_{n}^{ec}$ parameters, the IPM image $\hat{I}$ (or IPM features $\hat{F}$, or IPM maps $\hat{M}$) can be obtained through a hypothetical height $h$:
\begin{equation}
   \hat{I} = Plane(Proj(X_i,Y_i),\ldots,Proj(X_h,Y_w)),
    \label{eq1}
\end{equation}
\begin{equation}
    Proj(X_i,Y_i) = \sum_n I_n(u,v), if (u<imH , v<imW),
    \label{eq2}
\end{equation}
\begin{equation}
    Z_c * \left[\begin{array}{l} u_o \\ v_o \\ 1\\ \end{array} \right] = T_{n}^{in} \cdot T_{n}^{ec} \cdot \left[\begin{array}{l} X_i \\ Y_i \\ h \\  \end{array} \right],
    \label{eq3}
\end{equation}
\begin{equation}
    \left[\begin{array}{l} u \\ v \\ 1\\ \end{array} \right] = \left[\begin{array}{lll} imW/W &0 &0 \\ 0 &imH/H &0 \\ 0 &0 &1\\ \end{array} \right] \cdot \left[\begin{array}{l} u_o \\ v_o \\ 1\\ \end{array} \right],
    \label{eq4}
\end{equation}
where $Plane$ represents the set of all grids. 
$Z_c$ is the depth value in the camera coordinate system. 
$u$ and $v$, $u_o$ and $v_o$ are the value in the pixel coordinate system.
$n$ is the number of cameras. 
$imH$ and $imW$ are the sizes of images $I_n$ (features or maps) in the perspective coordinate system. 
$H$ and $W$ are the sizes of the original images. 

\subsection{Proposed Pipeline of GenMapping}
As shown in Fig.~\ref{fig.frame}, the framework of GenMapping follows a triadic synergy structure, comprising one principal and two auxiliary components. 
The principal branch (Sec.~\ref{p-branch}) learns the global semantic features in IPM images. Synchronously, the dense perspective branch (Sec.~\ref{dp-branch}) focuses on understanding spatial relationships of features from perspective views. The sparse prior branch (Sec.~\ref{sp-branch}) relies on the latent drivable area knowledge from OpenStreetMap (OSM). 
Ultimately, the auxiliary branch performs feature alignment and fusion with the principal branch in the Triple-Enhanced Merging (Tri-EM) (Sec.~\ref{tri}). Additionally, we propose a Cross-View Map Learning (CVML) (Sec.~\ref{cvml}) to improve the joint learning capability and a Bidirectional Data Augmentation (BiDA) (Sec.~\ref{da}) to mitigate overfitting in training. 
The framework, fundamentally guided by semantic maps, can be flexibly integrated into other models, such as vectorized mapping models. 
In this paper, the input features from the semantic head are used as simple BEV features to be incorporated into vectorized mapping models.

\subsubsection{The Principal Branch}
\label{p-branch}
The input of this branch is the IPM image, $\hat{I}$, which is converted from the initial multi-view perspective images $I_n$, as explained in Eq.~\ref{eq1} to Eq.~\ref{eq4}. 
Note that $imH$ and $imW$ in Eq.~\ref{eq4} are the size of $I_n$. 
Yet, learning from IPM images faces the challenge of local geometric distortions. 
We consider whether it is possible to mitigate local distortions with a global strategy, such as the modern State Space Model (SSM), excelling at global mutual modeling and linear computational complexity, as illustrated in the works~\cite{vmamba,vmunet}. Therefore, we propose this branch with the help of long-distance dependencies captured from SSM.
The principal branch is an encoder-decoder construct based on the UNet architecture, consisting of individual Vision Mamba (VM) blocks. 
Concretely, the features fused from the encoder $F_{en}$ and the auxiliary branch are fed into the decoder to obtain the decoded features $F_{de}$. Finally, a conventional layer is a head to decode a semantic map $M_{bev}$. 

A VM block is composed of several Visual State Space (VSS) sub-blocks with two branches. In a VSS sub-block, before entering the two branches, a layer normalization function is used for the input $F_1{=}\mathrm{LN}(F_i)$. 
The first branch contains a linear layer ($\mathrm{Linear}$) and an activation function ($\mathrm{SiLU}$~\cite{silu}): 
\begin{equation}
    F_2 = \mathrm{SiLU}(\mathrm{Linear}(F_1)).
    \vspace{-0.5em}
\end{equation}

In the second branch, the features successively pass through the linear layer ($\mathrm{Linear}$), depthwise separable convolution ($\mathrm{DSConv}$), the activation function ($\mathrm{SiLU}$) and a 2D-Selective-Scan module ($\mathrm{SS2D}$), as Eq.~\ref{e5}.
\begin{equation}
\label{e5}
    F_3 =\mathrm{SiLU} (\mathrm{DSConv}(\mathrm{Linear}(F_1))),
\end{equation}
\begin{equation}
    F_4 = \mathrm{SS2D}(F_3).
\end{equation}

{Moreover, the $\mathrm{SS2D}$ has three components: a scan expanding operation, an S6 block, and a scan merging operation, which is similar to VMamba~\cite{vmamba}.
After a Layer Normalization ($\mathrm{LN}$), $F_4$ is fused with the first by an Element-wise Production ($\mathrm{EP}$).
Then, the fused features $F_{fuse}$ learned through a linear layer are combined with a residual connection to output $F_o$. 
\begin{equation}
\label{e7}
    F_{fuse} = \mathrm{Linear}(\mathrm{EP}(\mathrm{LN}(F_4),F_2)),
    \vspace{-0.5em}
\end{equation}
\begin{equation}
    F_o = F_{fuse} + F_i.
    \vspace{-0.5em}
\end{equation}

\subsubsection{The Dense Perspective Branch}
\label{dp-branch}
Given that IPM images capture only the road plane features, information above the road plane is lost. 
This branch aims to supplement different information derived from perspective images, which are considered from two aspects. 
First, while the visual description of the road in IPM images and perspective images are similar, the surrounding distribution of the same structures appears differently in the two images due to differing coordinate systems, owning differentiated local feature distributions.
Additionally, IPM images only retain the road plane from perspective images, lacking interactions with other dynamic and static objects above the road plane, as shown in Fig.~\ref{fig.frame}.
These interactions, however, can be thoroughly explored in perspective images. 
Therefore, multi-view perspective images in this branch are fed into a lightweight semantic segmentation network to capture rich road features. 

In this section, we aim to exploit differentiated local features of road structures in perspective images. 
A classic lightweight convolutional network, ERFNet~\cite{erfnet}, is chosen. 
It balances the relationship between accuracy and efficiency through the design of non-bottleneck-1D modules, enabling efficient capture of contextual information.
\begin{equation}
   F_{pv} = \mathrm{E_{p}}(\sum_{i=0}^{n} I_i), 
\end{equation}
\begin{equation}
   M_{pv} = \mathrm{D_{p}}(F_{pv}).
\end{equation}

Finally, the road map $M_{pv}$ on the perspective images and the deep features of perspective $F_{pv}$ are obtained.

\subsubsection{The Sparse Prior Branch}
\label{sp-branch}
In simple environments, IPM images can accurately depict road planes. 
However, in complex scenarios, IPM images may suffer from severe spatial distortion issues that hinder accurate road structure localization, as shown in Fig.~\ref{fig.frame}. 
Therefore, in this section, we address these issues by leveraging sparse prior knowledge from OpenStreetMap (OSM). It describes the centerline of the driveable area in vector form. 

GPS coordinates from the vehicle can assist in capturing OSM within a specified range from the database. 
Since OSM data is the vector format, each local OSM data can be rasterized to obtain a grid map representation of OSM, $M_o{\in}\mathbb{R}^{1\times h\times w}$, which is used as the input for this branch. 
To maintain the shape of the principal branch, the padding operation $\mathrm{Pad}$ is used first:
\begin{equation}
    M_{o}^{'} = \mathrm{Pad}(M_o).
\end{equation}
Then, two unit layers composed of the convolution layer are designed to obtain the recessive feature of the drivable region.
\begin{equation}
    F_{o} = \mathrm{L_{d4}}(\mathrm{L_{d8}}(M_{o}^{'})),
\end{equation}
\begin{equation}
    \mathrm{L_{d8}} = (Repeat_3(C_{421},\mathrm{RELU}),\mathrm{BN}), 
\end{equation}
\begin{equation}
    \mathrm{L_{d4}} = (Repeat_2(C_{421},\mathrm{RELU}),C_{311},\mathrm{RELU},\mathrm{BN}),
\end{equation}
where $\mathrm{L_{d8}}$ and $\mathrm{L_{d4}}$ are all downsampling functions based on convolutional constructs. Here, $C_{421}$ represents that the convolution kernel is $4$, the stride is $2$, and the padding is $1$.
The definition of $C_{311}$ follows a similar pattern. 
$\mathrm{RELU}$ is an activation function. $\mathrm{BN}$ is a Batch Normalization operation. 

\begin{figure}[tb]
      \centering
      \includegraphics[scale=0.35]{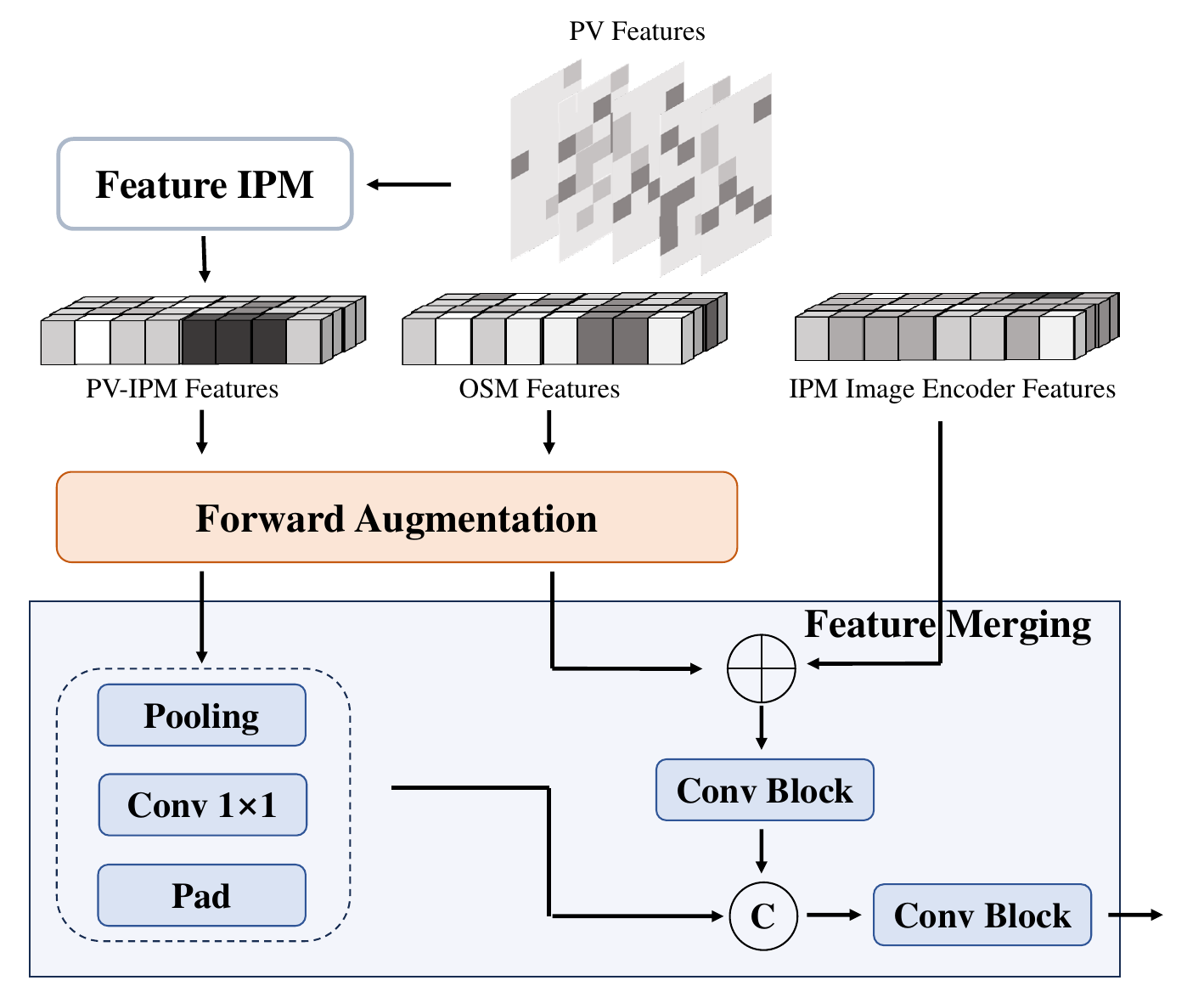}
      \caption{The details of the triple-enhanced merging module. `Conv Block' means `Convolutional Block', that is, $\mathrm{CB}$ in Sec.~\ref{tri}.}
      \label{fig.trifusion}
      \vspace{-1.0em}
\end{figure}
\subsubsection{Triple-Enhanced Merging}
\label{tri}
After synchronizing learning with two auxiliary branches, perspective features $F_{pv}$ can be obtained from the dense perspective branch, whereas OSM features $F_{o}$ are received from the sparse prior branch.
These auxiliary features are aggregated in this module between the encoder and decoder of the principal branch, as shown in Fig.~\ref{fig.trifusion}.

Since perspective features are not in the BEV space, perspective features in BEV coordinate system $\hat{F}_{pv}$ are obtained through the feature IPM technique, as explained in Eq.~\ref{eq1} to Eq.~\ref{eq4}. Here, $I_n$ is replaced by the perspective feature $F_{pv}$ in the formula, and other parameters are adjusted accordingly.
Note that the resolution of $\hat{F}_{pv}$ is lower than that of the other two branches, as a high sampling resolution in IPM can lead to information loss.
Then, the auxiliary features in the BEV space, $F_{o}$, $\hat{F}_{pv}$, jointly perform forward data augmentation, which will be discussed in Sec.~\ref{da}. 

The multi-branch features after the same data augmentation will be gradually fused. 
First, given deep features $F_{en}$ of the principal branch and $F_{o}$ of the prior branch, we directly add them. And a convolutional block $\mathrm{CB}$, consisting of a 1D convolution layer, an activation function, and a normalization layer, is applied to obtain a shallowly merged feature $F_{ms}$, as Eq.~\ref{eqa}:
\begin{equation}
\label{eqa}
    F_{ms} = \mathrm{CB}(\mathrm{Add}(F_{en},F_{o})).
\end{equation}
Next, the features $\hat{F}_{pv}$ are supplemented onto the shallow fusion features. 
Since the shape of $\hat{F}_{pv}$ is different with the shallow fusion features $F_{ms}$, we use pooling and convolution operations $\mathrm{L_d}$ on $\hat{F}_{pv}$, as Eq.~\ref{eqb}.
\begin{equation}
\label{eqb}
\begin{aligned}
    \hat{F}_{pv}^{'} &= \mathrm{L_d}(\hat{F}_{pv}).
\end{aligned}
\end{equation}
Before concatenating with the shallow fusion features $F_{ms}$, $\hat{F}_{pv}^{'}$ are again refined with padding $Pad$ to ensure shape consistency. 
Finally, a convolutional block $\mathrm{CB}$ is further used on the concatenated features. The enhanced feature $F_{me}$ can be obtained:
\begin{equation}
    F_{me} = \mathrm{CB}([F_{ms},\mathrm{Pad}(\hat{F}_{pv}^{'})]).
\end{equation}
After triple-enhancement fusion, $F_{me}$ serves as the input to the decoder module of the principal branch.

\subsection{Cross-view Map Learning}
\label{cvml}
Compared to map construction in the BEV space, road mapping in the perspective view tends to produce more robust construction results. 
This is because the perspective view directly employs the raw and accurate sensor data, whereas the BEV inevitably introduces uncertainty due to the view transformation. 
In other words, the robust semantic mapping in the perspective view can also serve as a joint learning signal to constrain BEV map construction, enhancing the generalization capacity of the model. Thus, this section proposes a map learning module in the common space between perspective view and BEV.

The perspective branch produces road maps $M_{pv}$, whereas the principal branch generates semantic maps $M_{bev}$ in the BEV space. 
To facilitate the establishment of mutual supervision across the global road structure with different semantics, both maps in the common space are described in terms of grid representations with binary. As the perspective map is already binary, we only convert $M_{bev}$ into binary maps.
\begin{equation}
    M_{bev}^{'} = \left\{\begin{array}{ll}
    0&grid=0, \\
	1 & else. \\
    \end{array}\right\}
\end{equation}

Similarly, it encounters the inconsistent issue of coordinate systems between two road maps.
Thus, an IPM-based approach is again employed. 
The IPM map $\hat{M}_{pv}$ translated from the perspective maps is obtained by Eq.~\ref{eq1} to Eq.~\ref{eq4}. 
$I_n$ is updated by $M_{pv}$, and $imH$ and $imW$ are the size of $M_{pv}$.
After obtaining the map in the same coordinate system, a loss is designed to constrain the model in terms of joint learning. It is defined as Eq.~\ref{ll}:
\begin{equation}
\label{ll}
    Loss_{jl} = L(\hat{M}_{pv}, M_{bev}^{'}),
\end{equation}
where $L$ denotes the L1 loss function.

\subsection{Bidirectional Data Augmentation}
\label{da}
Data augmentation is one of the effective techniques for enhancing generalization. 
Currently, in BEV map research~\cite{maptr,streammapnet}, data augmentation methods are applied in the perspective view, with very few methods available for augmentation in the BEV space. 
It is difficult to obtain accurate BEV features learned from scratch in these methods if additional uncertainties appear with data augmentation. 
In contrast, the BEV features in this method are directly based on IPM road images, IPM perspective features, and OSM features, none of which are learned starting from scratch. 
Therefore, different from existing data augmentation methods, a bidirectional data augmentation module is proposed in the BEV space.

Bidirectional data augmentation includes forward augmentation in the main pipeline and backward augmentation for extending mapping tasks.
The forward data augmentation faces three kinds of data: IPM road images, perspective IPM features, and OSM features. 
To ensure alignment in different kinds, geometric operations are selected, \textit{e.g.}, rotation and flipping, and both work together. 
Backward data augmentation is applied in other mapping tasks, such as vectorized mapping.
With the BEV features extracted from the pipeline, a misalignment between the features and the truth labels of the vectorized mapping task arises. 
To resolve this problem, we opt for inverse data augmentation methods for the processed features, employing secondary data augmentation to further decrease data dependence. Importantly, it is also better suited as a plug-and-play data augmentation method for extension to other tasks.

\begin{table*}[t]
\caption{Results of semantic mapping on nuScenes and Argoverse datasets. IoU is used as the evaluation metric.}
\vspace{-0.5em}
\label{sem-full}
\large
\centering
\fontsize{9}{12.8}\selectfont
\renewcommand{\arraystretch}{1.1}
\setlength\tabcolsep{7pt}
\begin{tabular}{lcc|cccc|cccc}
\toprule [2pt]
\multirow{2}{*}{Method} & \multirow{2}{*}{View Transformer} & \multirow{2}{*}{OSM} & \multicolumn{4}{c|}{nuScenes (IoU)}   & \multicolumn{4}{c}{Argoverse (IoU)}  \\ \cline{4-11} 
  &  &   & Div  & Ped  & Bou  & All Class & Div  & Ped  & Bou  & All Class \\ \hline
IPM-a~\cite{hdmapnet}   & IPM  &  & 14.4 & 9.5  & 18.4 & 14.1      & -    & -    & -    & -         \\
IPM-b~\cite{hdmapnet}   & IPM  &  & 25.5 & 12.1 & 27.1 & 21.6      & -    & -    & -    & -         \\
IPM-c~\cite{hdmapnet}   & IPM  &  & 38.6 & 19.3 & 39.3 & 32.4      & -    & -    & -    & -         \\
LSS~\cite{LSS}     & Depth  &  & 39.5 & 15.5 & 40.7 & 31.9    & 34.1 & 5.5  & 26.2 & 36.9      \\
HDMapNet~\cite{hdmapnet} & MLP   &   & 42.1 & 21.1 & 42.8 & 35.3   & 57.3 & 28.1 & 47.3 & 44.2      \\
BEVFormer~\cite{BEVFormer}   & BEVFormer  &  & 42.1 & 23.8 & 41.6 & 35.8      & -    & -    & -    & -         \\
P-MapNet~\cite{pmapnet}  & MLP   & \checkmark  & 44.1 & 22.6 & 43.8 & 36.8 & 52.9 & 29.7 & 46.8 & 43.1    \\
\rowcolor[HTML]{EFEFEF} 
GenMapping  & IPM   & \checkmark   & \textbf{46.1} & \textbf{30.5} &\textbf{44.5}  & \textbf{40.4}     & \textbf{59.3} & \textbf{37.0} & \textbf{48.4} & \textbf{49.1}   \\ \bottomrule [2pt]
\end{tabular}
\vspace{+1em}

\caption{Results of vectorized mapping on nuScenes and Argoverse. AP is used as the metric.}
\vspace{-0.5em}
\fontsize{9}{13.8}\selectfont
\label{vec_nus_a}
\begin{tabular}{c|cc|cccc|cccc}
\toprule [2pt]
\multicolumn{1}{c|}{\multirow{2}{*}{Strategy}} & \multirow{2}{*}{Method}  & \multirow{2}{*}{Image Size} & \multicolumn{4}{c|}{nuScenes (AP)} & \multicolumn{4}{c}{Argoverse (AP)} \\ \cline{4-11} 
\multicolumn{1}{c|}{} &    &  & Div & Ped & Bou & All Class & Div & Ped & Bou & All Class \\ \hline
\multirow{8}{*}{Normal} & VectorMapNet~\cite{vecmapnet}  & 455${\times}$256 & 47.3 & 36.1 & 39.3 & 40.9 & 36.1 & 38.3 & 39.2 & 37.9 \\
& InstaGraM~\cite{instmap}  & -- & 47.2 & 33.8 & 44.0 & 41.7 & - & - & - & - \\
& MapTR~\cite{maptr}  & 800${\times}$450 & 51.5 & 46.3 & 53.1 & 50.3 & 62.7 & 55.0 & 58.5 & 58.8 \\
& MapVR~\cite{vect2raster}  & 800${\times}$450 & 54.4 & 47.7 & 51.4 & 51.2 & 60.0 & 54.6 & 58.0 & 57.5 \\
& PivotNet~\cite{pivotnet}  & -- & 56.5 & 56.2 & 60.1 & 57.6 & - & - & - & - \\
& BeMapNet~\cite{bemapnet}  & 896${\times}$512& 62.3 & 57.7 & 59.4 & 59.8 & - & - & - & - \\
& MapTRv2~\cite{maptrv2}  & 800${\times}$450 & 61.4 & 57.8 & 60.4 & 59.9 & \textbf{68.8} & 61.3 & \textbf{63.4} & \textbf{64.5} \\
&\cellcolor[HTML]{EFEFEF}GenMapping  &\cellcolor[HTML]{EFEFEF}352${\times}$128 &\cellcolor[HTML]{EFEFEF}\textbf{63.7} &\cellcolor[HTML]{EFEFEF}61.5 &\cellcolor[HTML]{EFEFEF}61.2 &\cellcolor[HTML]{EFEFEF}62.1    &\cellcolor[HTML]{EFEFEF}64.3  &\cellcolor[HTML]{EFEFEF}52.3  &\cellcolor[HTML]{EFEFEF}56.3  &\cellcolor[HTML]{EFEFEF}57.8  \\ \hline
\multicolumn{1}{c|}{\multirow{2}{*}{Stream}} 
& StreamMapNet~\cite{streammapnet}  & 800${\times}$480 & 63.4 & 58.5 & 59.2 & 60.3 & - & - & - & 57.7 \\
\multicolumn{1}{c|}{} 
& \cellcolor[HTML]{EFEFEF}GenMapping  & \cellcolor[HTML]{EFEFEF}352${\times}$128 &\cellcolor[HTML]{EFEFEF}62.7 &\cellcolor[HTML]{EFEFEF}\textbf{63.9} &\cellcolor[HTML]{EFEFEF}\textbf{63.2} &\cellcolor[HTML]{EFEFEF}\textbf{63.2} &\cellcolor[HTML]{EFEFEF}55.0 &\cellcolor[HTML]{EFEFEF}\textbf{62.9} &\cellcolor[HTML]{EFEFEF}60.3 &\cellcolor[HTML]{EFEFEF}59.4
  \\ \bottomrule [2pt]
\end{tabular}
\vspace{-1.0em}
\end{table*}
\subsection{Loss Function}
For the supervision of the proposed model, the whole training loss is composed of four parts: semantic mapping loss $Loss_{hd}$, perspective mapping loss $Loss_{pv}$, joint learning loss $Loss_{jl}$, and additional task loss $Loss_{task}$:
\begin{equation}
    Loss = \alpha_1 \times Loss_{hd} + \alpha_2 \times Loss_{pv} + \alpha_3 \times Loss_{jl} + Loss_{task},
\end{equation}
where weight relationship ${\alpha}_3{=}0.1{\times}{\alpha}_1$ is applied to each task. 
$Loss_{hd}$ and $Loss_{pv}$ are cross-entropy loss functions. 

This paper further explores the task of vectorized mapping, where $Loss_{task}$ follows the settings of the reference paper, and the proposed weights such as ${\alpha}_1$, ${\alpha}_2$, and ${\alpha}_3$ will be adjusted.

\section{Experiment}
\begin{figure*}[htb]
      \centering
      \includegraphics[scale=0.23]{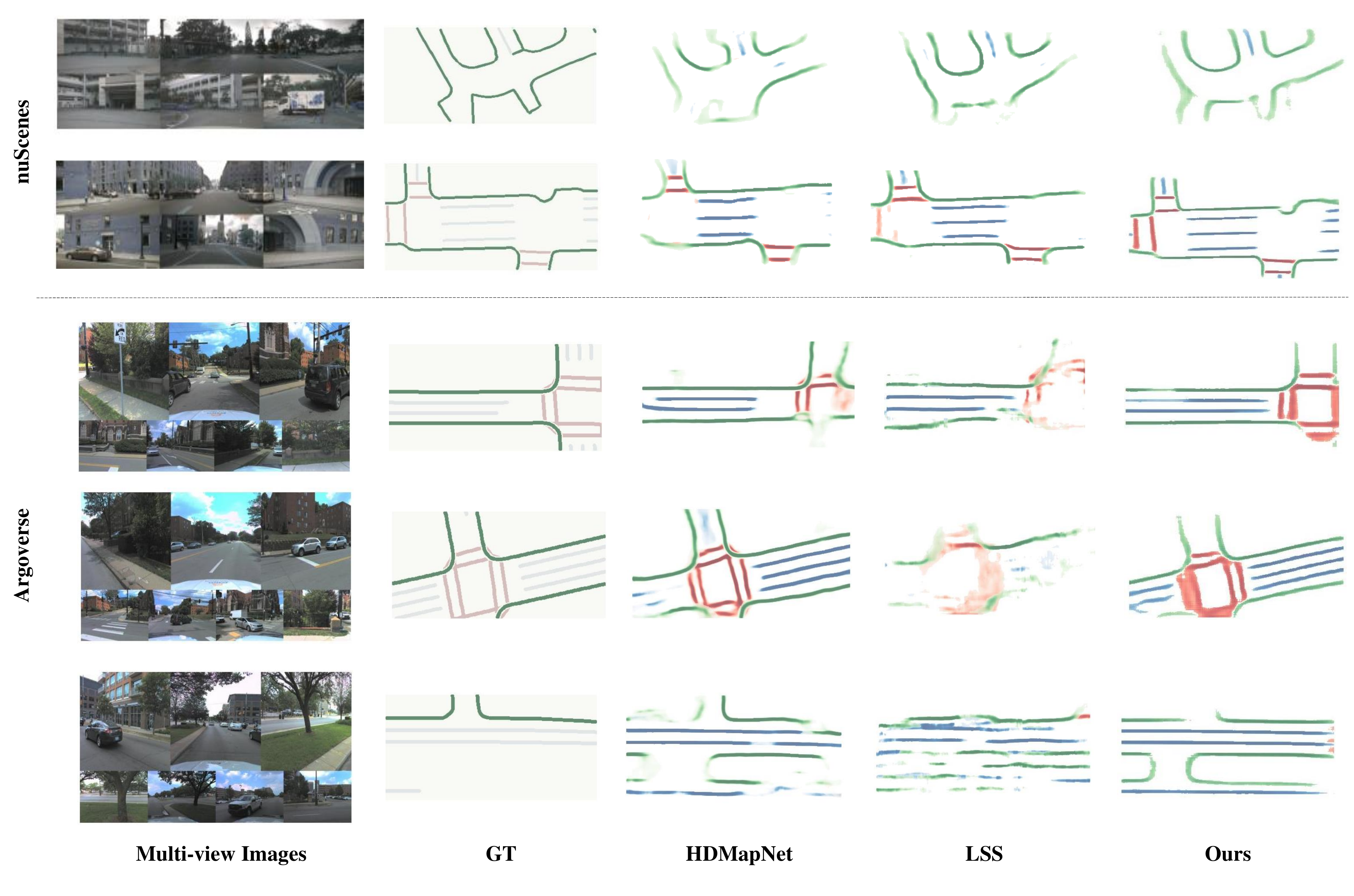}
      \caption{Visualization results for semantic mapping. The proposed method is compared against state-of-the-art semantic mapping methods including HDMapNet~\cite{hdmapnet} and LSS~\cite{LSS}.  Classes of divider, pedestrian, and boundary are filled with green, red, and blue.}
      \label{fig.sem}
      \vspace {-1.5em}
\end{figure*}

\subsection{Datasets and Metrics}
\subsubsection{Datasets}
We evaluate our method on two widely used datasets in HD maps construction, nuScenes~\cite{nus} and Argoverse~\cite{argoverse}.
The nuScenes dataset, collected in Singapore and Boston, includes six cameras, a 32-beam LiDAR, five radars, and their respective internal and external parameters. 
It also provides GPS location data, which can be used to derive the corresponding OpenStreetMap (OSM) map.
However, the original nuScenes contains overlapping locations in the train and validation sets, leading to potential model memorization of existing map structures.
Thus, the work~\cite{streammapnet} repartitioned the sets to construct a new-split nuScenes dataset.
In the Argoverse dataset, seven cameras capture RGB images at $20$Hz and the LiDAR data contains $20000$ sequences. Similarly, each scene also has an HD map that includes 3D lanes and sidewalks.

Based on experimental requirements, this section divides the datasets with two splitting settings.
The first follows the setting used in most HD map construction models: the training and validation sets are split into $700$ and $150$ scenes in the nuScenes dataset, whereas the Argoverse dataset is also divided according to the official settings.
The second setting focuses on verifying the generalization capacity. 
Experiments are conducted on a non-overlapping new-split nuScenes dataset. 
Then, cross-dataset experiments are carried out between nuScenes and Argoverse datasets, with each dataset serving alternately as the training and validation set. 
Finally, cross-location validation is performed within the nuScenes dataset, where the training and validation sets are split based on different cities.

\subsubsection{Metrics}
Conventionally, three map elements, covering lane divider, pedestrian crossing, and road boundary, are selected for evaluation. 
For semantic mapping, we adopt the Intersection over Union (IoU) metric as per the standard in~\cite{hdmapnet}. 
For vectorized mapping, we use Average Precision (AP), which is calculated based on Chamfer Distance (CD) thresholds of $\{0.5m,1.0m,1.5m\}$.
All class metric is obtained by averaging the value of three classes. 
For testing and analyzing the generalization performance, the generalization ratio is employed:
\begin{equation}
    Ratio = M_{A2B} / M_A,
\end{equation}
where $M_{A2B}$ refers to the results of the model trained on $A$ dataset evaluated on $B$ dataset. $M_A$ is the results of testing on $A$ dataset by the same model. 

\subsection{Implementation Details}
All experiments are conducted with an NVIDIA RTX A6000 GPU. 
In addition, the experiment results are obtained using only visual sensor data as input. The range of an HD map and an OSM map is $({-}30m,30m)$ on the X-axis and $({-}15m,15m)$ on the Y-axis, respectively. Besides, the map resolution is $0.15m$. 

\textbf{Semantic HD mapping:} 
The model is trained on the nuScenes dataset for $30$ epochs and the Argoverse dataset for $8$ epochs. 
The parameters ${\alpha}_1$, ${\alpha}_2$ and ${\alpha}_3$ are set to $1$, $1$ and $0.1$, respectively.  
The batch size is $8$ and the initial learning rate is $2.5e^{-4}$. 
AdamW is adopted as the optimizer. 
CosineAnnealingLR is employed as the learning strategy with $500$ iterations and a minimum learning rate of $1e^{-5}$.

\textbf{Vectorized HD mapping:} 
With a batch size of $8$, the initial learning rate for the view transformer module is set to $2.5e^{-4}$. 
The decoder module adheres to reference works~\cite{maptrv2,streammapnet}, using $1.5e^{-4}$ for the normal strategy and $1.25e^{-4}$ for the stream strategy. 
The training epochs, optimizer, and learning schedule are also consistent with those in the reference works.

\begin{figure*}[htb]
      \centering
      \includegraphics[scale=0.25]{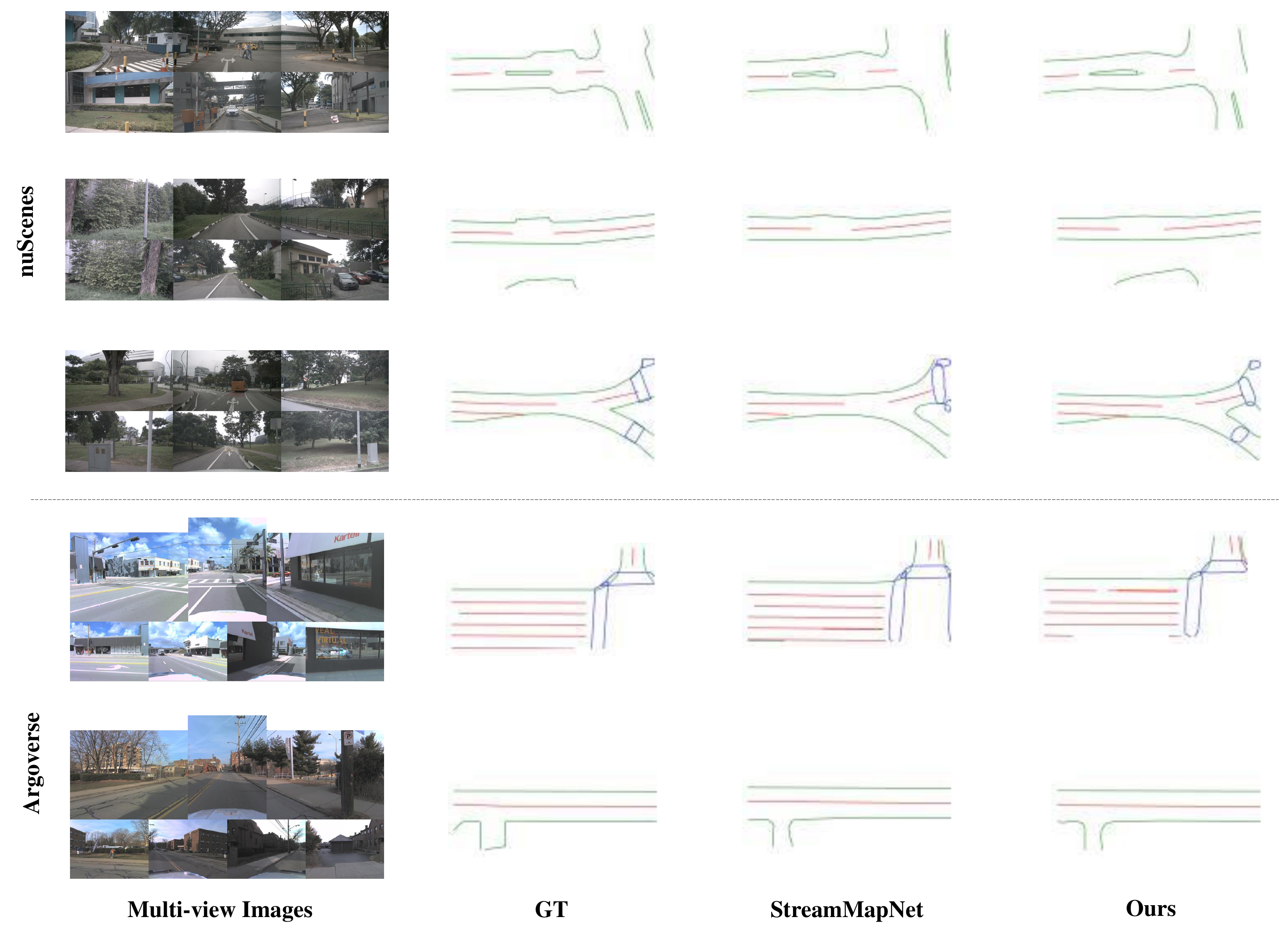}
      \caption{Visualization results for vectorized mapping. The proposed method is compared against a state-of-the-art vectorized mapping method including StreamMapNet~\cite{streammapnet}. Classes of divider, pedestrian, and boundary are filled with green, blue, and red.}
      \label{fig.vec}
      \vspace {-1.5em}
\end{figure*}

\subsection{Comparison with State-of-the-Art Methods}
\textbf{Semantic HD mapping:}
We choose two competitive semantic mapping methods for comparison: HDMapNet~\cite{hdmapnet} and P-MapNet~\cite{pmapnet}. 
In addition, we compare the mapping capabilities of other view transformation modules, \textit{i.e.}, LSS~\cite{LSS} and BEVFormer~\cite{BEVFormer} which are completed on the framework of~\cite{hdmapnet}. 
IPM-a, IPM-b, and IPM-b are different designs of IPM, derived from the work of~\cite{hdmapnet}.
As shown in Table~\ref{sem-full}, GenMapping outperforms existing approaches under different datasets by a significant margin. 
The proposed method achieves $40.4\%$ in mIoU on the nuScenes dataset and $49.1\%$ in mIoU on the Argoverse dataset, yielding respective ${+}3.6\%$ and ${+}6.0\%$ gains. 
It becomes clear that our method has outstanding performance in semantic mapping. 
The visualization results in Fig.~\ref{fig.sem} further corroborate that our method provides precise details in map structure compared to others.

\textbf{Vectorized HD mapping:} 
According to the current research on vectorized mapping, it can be roughly divided into non-temporal works (normal) and temporal merging works (stream). 
In the non-temporal works, MapTR~\cite{maptr} and MapTRv2~\cite{maptrv2} are representative works. 
The former takes GKT~\cite{gkt} as the view transformation module, whereas the latter employs BEVPool~\cite{bevpoolv2} with depth ground truth. 
In addition, we compare our work with other relevant studies.
The temporal works, exemplified by StreamMapNet~\cite{streammapnet}, use a streaming strategy to fuse temporal features. 
This method adopts BEVFormer~\cite{BEVFormer} as the view transformer module. 
Therefore, we validate the proposed method in two parts, as shown in Table~\ref{vec_nus_a}.  
In both strategies, GenMapping is used as the view transformer. 
The normal strategy employs the same decoding method as MapTRv2, whereas the stream strategy utilizes the decoder from StreamMapNet. 
The results demonstrate that the proposed method has outstanding performance, surpassing the baseline models with higher scores of $63.2\%$ and $59.4\%$ in mAP. 
Moreover, this highlights that GenMapping can be seamlessly integrated into vectorized mapping tasks, showcasing its plug-and-play capability. 
Fig.~\ref{fig.vec} illustrates the visualization results of vectorized mapping, where the proposed method provides more comprehensive instance detection, particularly for subtle road structures such as distant crosswalks. 

\begin{table}[t]
\caption{Results of semantic mapping on the new-split nuScenes dataset.}
\vspace{-1.0em}
\label{sem_new}
\fontsize{8}{12.8}\selectfont
\begin{center}
\resizebox{0.45\textwidth}{!}{
\begin{tabular}{l|cccc}
\toprule [2pt]
\multirow{2}{*}{Method} & \multicolumn{4}{c}{Result (IoU)}  \\ \cline{2-5}
& Div & Ped & Bou & All Class \\ \hline
LSS~\cite{LSS}  & 27.2 & 10.7 & 29.1 & 22.3 \\
HDMapNet~\cite{hdmapnet}   & 26.1 & 13.8 & 26.8 & 22.2 \\
P-MapNet~\cite{pmapnet}  & 26.2 & 5.0 & 24.9 & 17.2 \\
\rowcolor[HTML]{EFEFEF}
GenMapping  & 28.8 & 19.7 & 26.4 & \textbf{25.0} \\ \bottomrule [2pt]
\end{tabular}}
\end{center}
\vspace{-2.4em}
\end{table}

\begin{table}[t]
\caption{Results of vectorized mapping on the new-split nuScenes dataset.}
\vspace{-0.5em}
\fontsize{8}{12.8}\selectfont
\centering
\label{vec_new}
\resizebox{0.45\textwidth}{!}{
\begin{tabular}{l|cccc}
\toprule [2pt]
\multirow{2}{*}{Method} & \multicolumn{4}{c}{Result (AP)} \\ \cline{2-5}
&Div & Ped & Bou & All Class \\ \hline
MapTR~\cite{maptr}  & 20.7 & 6.4 & 35.5 & 20.9 \\
MapTRv2~\cite{maptrv2}  & 24.8 & 13.0 & 42.4 & 26.7 \\
StreamMapNet~\cite{streammapnet}  & 30.2 & 27.5 & 38.1 & 31.9 \\
\rowcolor[HTML]{EFEFEF}
GenMapping & 28.9 & 35.4 & 38.5 & \textbf{34.3} \\ \bottomrule [2pt]
\end{tabular}}
\vspace{-2.0em}
\end{table}

\subsection{New-Split Dataset Experiments}
Considering that the original split of the nuScenes dataset involves scene overlap, the work~\cite{streammapnet} proposes a new-split dataset to assess model generalization. 
We evaluate all mapping methods on this new split, all of which significantly decline in accuracy, indicating that generalization research is highly necessary.
As shown in Table~\ref{sem_new}, the $25\%$ in mIoU achieved on a dataset with none-overlap demonstrates the generalization capability of our method. 
Additionally, the effectiveness of vectorized mapping is shown in Table~\ref{vec_new}. 
As observed, our method achieves the highest accuracy, reaching $34.3\%$ in mAP.
It can be confidently thought that the proposed method seamlessly integrates into vectorized mapping tasks and maintains high generalization efficiency.

\begin{table}[t]
\caption{Cross-dataset validation of semantic mapping. `Nus' denotes nuScenes and `Arg' denotes Argoverse.}
\centering
\large
\label{sem_cross}
\fontsize{7}{11.8}\selectfont
\resizebox{0.45\textwidth}{!}{
\begin{tabular}{l|c|c|cc}
\toprule [2pt]
Method & Train & Val & mIoU & Ratio \\ \hline
\multirow{2}{*}{LSS~\cite{LSS}} & \multirow{6}{*}{Nus} & Nus & 31.9 & \multirow{2}{*}{5.6} \\
 &  & Arg & 1.78 &  \\ \cline{1-1} \cline{3-5}
\multirow{2}{*}{HDMapNet~\cite{hdmapnet}} &  & Nus & 35.3 & \multirow{2}{*}{10.5} \\
 &  & Arg & 3.71 &  \\ \cline{1-1} \cline{3-5}
\multirow{2}{*}{GenMapping} &  & Nus & 40.4 & \multirow{2}{*}{\textbf{25.1}} \\ 
 &  & Arg & 10.1 &  \\ \bottomrule [2pt]
\multirow{2}{*}{LSS~\cite{LSS}} & \multirow{6}{*}{Arg} & Arg & 36.9 & \multirow{2}{*}{3.0} \\
 &  & Nus & 1.1 &  \\ \cline{1-1} \cline{3-5}
\multirow{2}{*}{HDMapNet~\cite{hdmapnet}} &  & Arg & 44.2 & \multirow{2}{*}{3.9} \\
 &  & Nus & 1.7 &  \\ \cline{1-1} \cline{3-5}
\multirow{2}{*}{GenMapping} &  & Arg & 49.1 & \multirow{2}{*}{\textbf{21.2}} \\
 &  & Nus & 10.4 & \\ \bottomrule [2pt]
\end{tabular}}
\vspace{-2.0em}
\end{table}
\begin{table}[t]
\caption{Cross-dataset validation of semantic mapping. `Nus' denotes nuScenes and `Arg' denotes Argoverse. `Nor' denotes the non-temporal strategy (normal) and `Str' denotes the temporal strategy (stream).}
\fontsize{9}{12.8}\selectfont
\centering
\label{vec_cross}
\resizebox{0.45\textwidth}{!}{
\begin{tabular}{l|c|c|c|cc}
\toprule [2pt]
 &Method & Train & Val & mAP & Ratio \\ \hline
\multirow{4}{*}{Nor}
&\multirow{2}{*}{MapTRv2~\cite{maptrv2}} & \multirow{8}{*}{Nus} & Nus & 59.9 & \multirow{2}{*}{0.0} \\
& &  & Arg & 0.0 &  \\ \cline{2-2} \cline{4-6}
& \multirow{2}{*}{GenMapping} & & Nus & 62.1 & \multirow{2}{*}{\textbf{7.2}} \\
& &  & Arg & 4.5 &  \\ \cline{1-2} \cline{4-6}
\multirow{4}{*}{Str}
&\multirow{2}{*}{StreamMapNet~\cite{streammapnet}} &  & Nus & 60.3 & \multirow{2}{*}{8.3} \\
& &  & Arg & 5.0 &  \\ \cline{2-2} \cline{4-6}
&\multirow{2}{*}{GenMapping} &  & Nus & 62.3 & \multirow{2}{*}{\textbf{12.8}} \\ 
& &  & Arg & 8.0 &  \\ \bottomrule [2pt]
\multirow{4}{*}{Nor}
& \multirow{2}{*}{MapTRv2~\cite{maptrv2}} & \multirow{8}{*}{Arg} & Arg & 64.5 & \multirow{2}{*}{0.0} \\
 & &  & Nus & 0.0 &  \\ \cline{2-2} \cline{4-6}
 & \multirow{2}{*}{GenMapping} &  & Arg & 57.8 & \multirow{2}{*}{\textbf{8.0}} \\
 & &  & Nus & 4.6 &  \\ \cline{1-2} \cline{4-6}
\multirow{4}{*}{Str}
& \multirow{2}{*}{StreamMapNet~\cite{streammapnet}} &  & Arg & 57.7 & \multirow{2}{*}{8.8} \\
&  &  & Nus & 5.1 &  \\ \cline{2-2} \cline{4-6}
& \multirow{2}{*}{GenMapping} &  & Arg & 59.4 & \multirow{2}{*}{\textbf{20.9}} \\
&  &  & Nus & 12.4 & \\ \bottomrule [2pt]
\end{tabular}}
\vspace{-1.8em}
\end{table}
\begin{figure}[tb]
      \centering
      \includegraphics[scale=0.45]{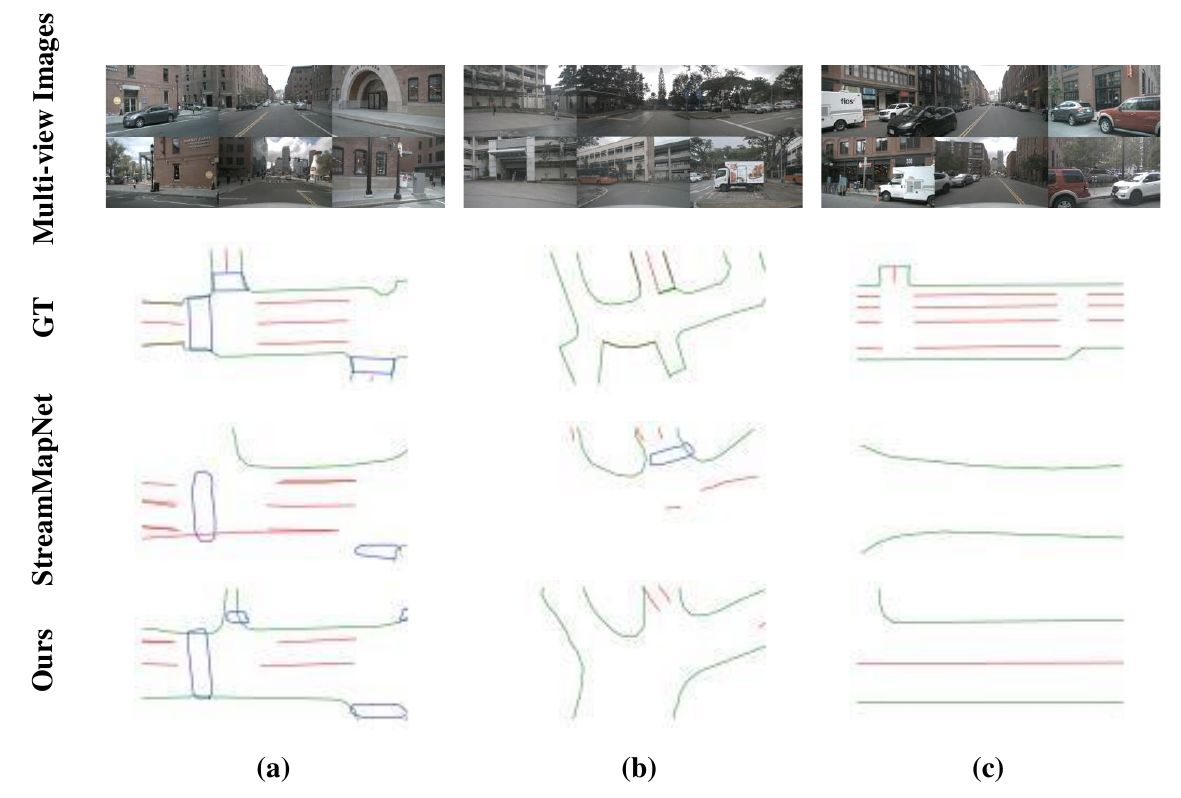}
      \vspace{-0.55cm}
      \caption{Visualization results of cross-dataset vectorized mapping. The model trained on the Argoverse dataset is verified on the nuScenes dataset.}
      \label{fig.vec_a2n}
      \vspace {-1.8em}
\end{figure}
\subsection{Cross-Dataset Experiments}
Table~\ref{sem_cross} presents the generalization validation of semantic mapping in two datasets where the sensor layout is not consistent.
The experiments successively replace nuScenes and Argoverse as training and validation sets, respectively. 
LSS~\cite{LSS} incorporates intrinsic parameters into learning depth, resulting in poor performance across datasets. 
Although HDMapNet~\cite{hdmapnet} decouples extrinsic parameters from model training, it still relies on model learning for intrinsic parameters. 
In contrast, our approach decouples both intrinsic and extrinsic parameters from training, offering better generalization performance compared to the previous two methods. 
The generalization ratios of the proposed method reach $25.1\%$ and $21.2\%$, improving by $14.9\%$ and $17.3\%$, respectively.

For vectorized mapping, Table~\ref{vec_cross} shows the generalization results across the two datasets. 
To ensure fairness, we validate under two strategies. 
Overall, the stream strategy with temporal fusion shows better generalization compared to the normal strategy. 
This is because the consistency of map instances across temporal sequences further helps constrain map construction. 
On closer inspection, the proposed method demonstrates stronger generalization performance in both strategies, achieving $12.8\%$ and $20.9\%$ ratios. 
Fig.~\ref{fig.vec_a2n} provides the cross-dataset visualization results of vectorized mapping.
As shown in Fig.~\ref{fig.vec_a2n}(a), in clear and common road environments, the maps generated by the proposed method are of higher quality. 
However, in complex road scenarios, such as those depicted in Fig.~\ref{fig.vec_a2n}(b) and (c), where perspective views often involve significant vehicle occlusions, the generalization performance is less satisfactory. This remains a challenge that needs to be addressed in future research.

\begin{table}[t]
\caption{Cross-location validation of semantic mapping.}
\centering
\large
\label{sem_cross_loc}
\fontsize{6}{9.0}\selectfont
\resizebox{0.45\textwidth}{!}{
\begin{tabular}{l|c|c|c}
\toprule [1.6pt]
Method & Train & Val & mIoU  \\ \hline
LSS~\cite{LSS} & \multirow{3}{*}{Boston} & \multirow{3}{*}{Singapore} &7.9   \\ 
P-MapNet~\cite{pmapnet} &  &  & 8.0  \\
GenMapping &  & &\textbf{9.7}     \\  \bottomrule [1.6pt]               
\end{tabular}}
\vspace{1.0 em}

\caption{Ablation result of core modules. `PB' is the principal branch. `FDA' means forward data augmentation.}
\centering
\large
\label{abl}
\fontsize{5}{7.0}\selectfont
\resizebox{0.45\textwidth}{!}{
\begin{tabular}{lcccc}
\toprule [1.5pt]
PB & Tri-EM & FDA & CVML & mIoU \\ \hline
 \checkmark  &  &  & &35.9 \\
 \checkmark  &\checkmark  & & & 38.0 \\
 \checkmark  &\checkmark  &\checkmark & & 39.1 \\
 \checkmark  &\checkmark  &\checkmark  &\checkmark &40.4\\ \bottomrule [1.5pt]
\end{tabular}}
\vspace{-1.0 em}
\end{table}
\subsection{Cross-Location Experiments}
In the nuScenes dataset, there are two places of data collection, \textit{i.e.}, Boston and Singapore. 
Given the differences in road environments and driving regulations between them, we assess cross-location generalization using consistent sensors. 
Table~\ref{sem_cross_loc} shows the cross-location results of semantic mapping. 
Overall, despite using the same sensor distribution, the cross-regional validation results are not particularly impressive. This issue may be due to a reduction in training data, leading to overfitting of the model. Nevertheless, our method still delivers an exceptional performance with an improvement of $1.7\%$ in mIoU. 

\subsection{Ablation Study}
In this section, we verify the effectiveness of the core modules, loss weights, and inference speed.

\begin{table}[t]

\caption{Ablation result of the principal branch. `N2A' means the cross-dataset IoU ratio of nuScenes to Argoverse.}
\centering
\large
\label{frame_pb}
\fontsize{8}{11.8}\selectfont
\resizebox{0.45\textwidth}{!}{
\begin{tabular}{l|cccc|c}
\toprule [2pt]
\multirow{2}{*}{PB} & \multicolumn{4}{c|}{Result (IoU)} & \multirow{2}{*}{N2A} \\ \cline{2-5}
 & Div & Ped & Bou & All class &  \\ \hline
ERFNet~\cite{erfnet} & 42.0 & 23.7 & 40.1 & 35.3 & 6.3 \\
UNetFormer~\cite{unetformer} & 38.4 & 23.0 & 37.2 & 32.9 & \textbf{11.0} \\
Mamba-UNet & \textbf{46.1} & \textbf{30.5} & \textbf{44.5} & \textbf{40.4} & 10.1 \\ \bottomrule [2pt]
\end{tabular}}
\vspace{1.0em}

\caption{Ablation result of Merging the OSM Branch.}
\centering
\large
\label{osm_fuse}
\fontsize{7}{10.8}\selectfont
\resizebox{0.45\textwidth}{!}{
\begin{tabular}{l|cccc}
\toprule [2pt]
\multirow{2}{*}{Method} & \multicolumn{4}{c}{Result (IoU)} \\ \cline{2-5}
 & Div & Ped & Bou & All class \\ \hline
Cross-attention & 43.8 & 25.9 & 42.0 & 37.4 \\
Add & \textbf{46.1} & \textbf{30.5} & \textbf{44.5} & \textbf{40.4} \\ \bottomrule [2pt]
\end{tabular}}
\vspace{-1.0em}
\end{table}
\begin{table}[t]
\caption{Analysis of the weight of loss in vectorized mapping tasks.}
\centering
\label{weight}
\fontsize{5}{6.0}\selectfont
\resizebox{0.45\textwidth}{!}{
\begin{tabular}{l|cccc}
\toprule [1.5pt]
Strategy & ${\alpha}_{1}$  &${\alpha}_{2}$ &${\alpha}_{3}$ & mAP \\ \hline
\multirow{3}{*}{Normal} 
& 1 & 1 & 0.1 & 58.4 \\
 & 5 & 5 & 0.5 & 59.4 \\
 & \textbf{10} & \textbf{10} & \textbf{1} & \textbf{62.1} \\ \hline
\multirow{2}{*}{Stream} & 5 & 5 & 0.5 & 61.1 \\
 & \textbf{10} & \textbf{10} & \textbf{1} & \textbf{63.2} \\ \bottomrule [1.5pt]
\end{tabular}}
\vspace{-2.0em}
\end{table}

\subsubsection{Effectiveness of Core Modules}
To validate the positive impact of each module in the design, we analyze the effectiveness of each module in the context of semantic mapping. 
Table~\ref{abl} shows the ablation results. 
The baseline is the principal branch. 
Then a triadic synergy framework with a triple-enhanced merging module is added to the baseline, reaching $38.0\%$ in mIoU with an improvement of ${+}2.1\%$. 
Next, the effect of forward data augmentation is demonstrated, which yields a gain of ${+}1.1\%$ in mIoU. 
Finally, the improvement of ${+}1.3\%$ in mIoU brought by CVML indicates that the map interaction between the perspective view and the BEV is feasible and advantageous to improve the quality of mapping.

\begin{figure*}[htb]
      \centering
      \includegraphics[scale=0.45]{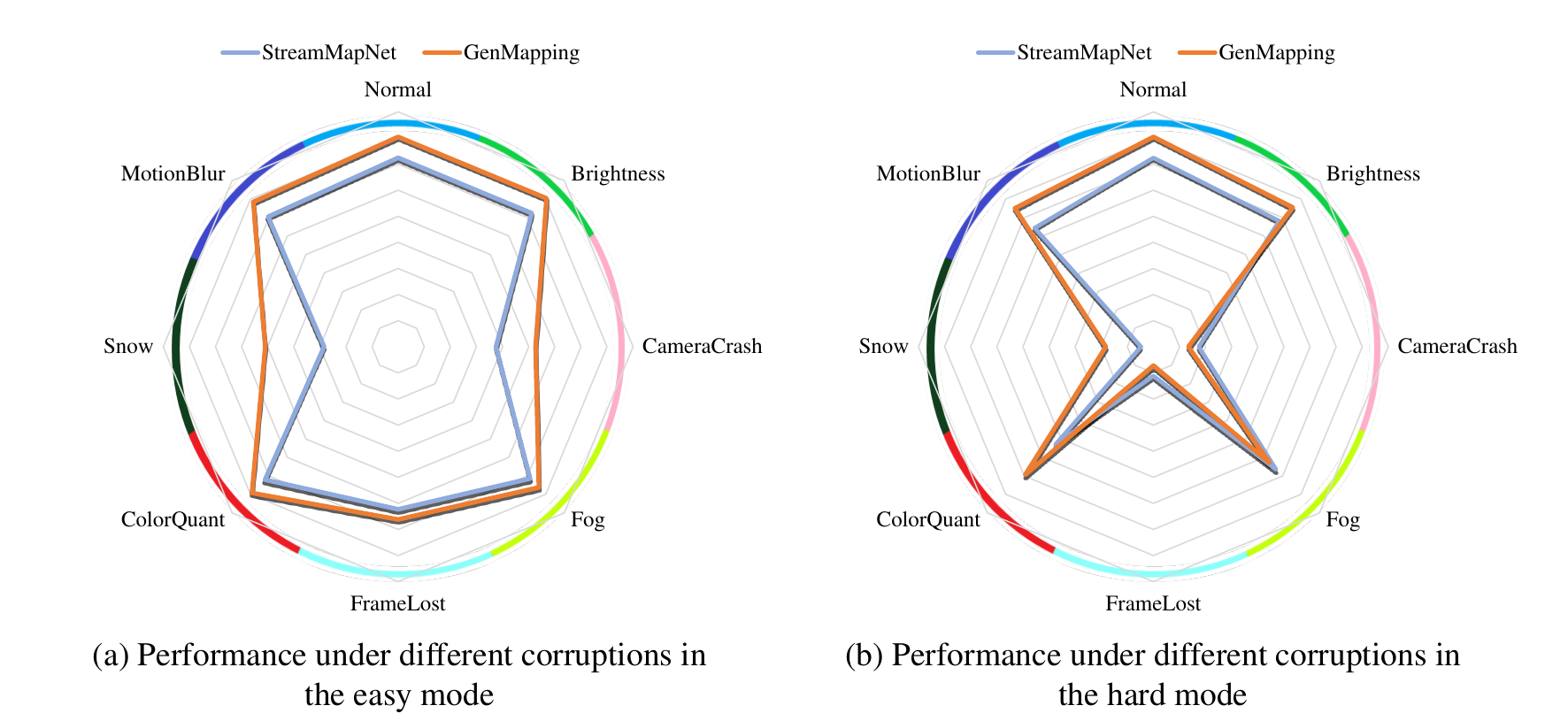}
      \vspace{-0.38cm}
      \caption{Performance analysis under different corruptions. The results denoted in orange are from our proposed GenMapping model, whereas the blue ones are from StreamMapNet~\cite{streammapnet}. The closer to the center, the lower the accuracy.}
      \label{fig.robotness}
      \vspace {-1.5em}
\end{figure*}

\subsubsection{Analysis of Basic Modules in the Principal Branch}
To further analyze the performance of Mamba architecture in HD mapping, we explore the performance of semantic mapping under different frameworks of the principal branch, as shown in Table~\ref{frame_pb}. 
ERFNet~\cite{erfnet} and UNetFormer~\cite{unetformer} are both encoder-decoder architectures. 
The former uses convolutional units as the basic blocks, while the latter employs the transformer. 
In cross-dataset generalization experiments, UNetformer demonstrates superior performance. 
Nonetheless, Mamba-UNet demonstrates comparable performance to UNetFormer in cross-data experiments, while delivering more excellent results on individual datasets. 
Thus, Mamba-UNet serves as the final choice in the framework. 

\subsubsection{Analysis of Merging OSM}
Due to OSM misalignment caused by GPS errors, there is an alignment fusion issue in the triple-enhanced merging module. 
In this section, we explore two ways to fuse the sparse OSM branch, as shown in Table~\ref{osm_fuse}. 
Due to the high feature resolution in this module, the misalignment effects have gradually diminished, whereas the direct addition method achieves higher learning efficiency.

\subsubsection{Analysis of Loss Weights in Vectorized Mapping Tasks}
In this section, we evaluate the weight relationship between the proposed loss and detection loss of other vectorized mapping tasks. Table~\ref{weight} shows the results of the experiment. 
Note that loss weights of detecting map instances are fixed and consistent with the reference paper in these experiments. 
It can be observed that when the weights are $10$, $10$, and $1$, the highest accuracy is achieved under both strategies. 

\subsubsection{Analysis of Efficiency}
In addition to map quality, models for online HD mapping also require fast inference speeds. 
Table~\ref{effi} presents the inference efficiency results of different models. Our method not only achieves higher accuracy but also faster inference speed. 
This rapid inference speed is attributed to both the source data and the model architecture. 
Concretely, the proposed online HD map construction method enables efficient usage of low-resolution images and incorporates a more lightweight state-space-model-based architecture, achieving a balance between efficiency and accuracy, which is perfectly suitable for real-world applications. 

\begin{table}[t]
\caption{Efficiency results of different methods.}
\centering
\large
\label{effi}
\fontsize{8}{11.8}\selectfont
\resizebox{0.48\textwidth}{!}{
\begin{tabular}{c|c|ccc}
\toprule [2pt]
Strategy & Metric & MapTR~\cite{maptr} & MapTRv2~\cite{maptrv2} & Ours \\ \hline 
\multirow{2}{*}{Normal} & mAP &50.3 &59.9 &\textbf{62.1} \\
 & FPS & 6.1 & 5.6 & \textbf{7.4} \\ \toprule [2pt]
Strategy & Metric & \multicolumn{2}{c}{StreamMapNet~\cite{streammapnet}} & Ours \\ \hline
\multirow{2}{*}{Stream} & mAP &\multicolumn{2}{c}{60.3} &\textbf{63.2} \\
 &FPS &\multicolumn{2}{c}{5.6} &\textbf{6.8} \\ \bottomrule[2pt]
\end{tabular}}
\vspace{-2.0em}
\end{table}
\subsection{Analysis of Robustness against Corruptions}
Sensor data is also a crucial factor affecting model quality and robustness. 
In this section, we assess the robustness of our model under different sensor corruptions. 
We utilize the nuScenes-C dataset proposed in the work~\cite{robobev} as our dataset benchmark. 
It involves data corruption performed on the validation set of nuScenes.  
Seven types of corruption are chosen to evaluate: Brightness, CameraCrash, Fog, FrameLost, ColorQuant, Snow, and MotionBlur.  
Fig.~\ref{fig.robotness} illustrates the robustness results under easy and hard corruptions.  It can be observed that our method demonstrates stronger robustness compared to the benchmark methods across a wide range of corruption scenarios, in particular under Snow, CameraCrash, and ColorQuant conditions.

\section{Conclusion}

After fully exploiting the potential of IPM, this paper proposes a generalizable map model, GenMapping. 
We design a triadic synergy framework, with IPM as the core for view transformation, effectively harnessing the advantages of parameter decoupling. 
Simultaneously, a cross-view map learning module and a bidirectional data augmentation module are introduced to further enhance the model's robustness and generalization. 
The state-of-the-art performance on nuScenes and Argoverse datasets demonstrates the versatility of the model in both semantic and vectorized mapping. 
In extensive experiments with identical sensors but non-overlapping datasets, as well as in cross-dataset and cross-region evaluation, the proposed method shows strong generalization capabilities.

There is still rich research space to further improve cross-dataset performance in complex environments. As discussed in the experiments, vehicle occlusion significantly impacts the quality of visual BEV mapping. 
Therefore, our future works will focus on developing effective strategies to mitigate the adverse effects of object occlusion. Additionally, the domain gap of BEV mapping across different datasets also warrants further investigation.

\bibliographystyle{IEEEtran}
\bibliography{ref.bib}

\end{document}